\documentclass{article}

\usepackage{amsmath}
\usepackage{amsthm}
\usepackage{bm}
\usepackage{amssymb}
\usepackage{microtype}
\usepackage{graphicx}
\usepackage{subfigure}
\usepackage{booktabs}
\usepackage{multirow}
\usepackage{amsmath}
\usepackage{bbding}
\usepackage{hyperref}

\usepackage[accepted]{icml2020}

\icmltitlerunning{Hierarchical Inter-Message Passing for Learning on Molecular Graphs}

\begin{document}

\twocolumn[
\icmltitle{Hierarchical Inter-Message Passing for Learning on Molecular Graphs}

\icmlsetsymbol{equal}{*}

\begin{icmlauthorlist}
\icmlauthor{Matthias Fey}{equal,tudo}
\icmlauthor{Jan-Gin Yuen}{equal,tudo}
\icmlauthor{Frank Weichert}{tudo}
\end{icmlauthorlist}

\icmlaffiliation{tudo}{Department of Computer Science, TU Dortmund University, Dortmund, Germany}

\icmlcorrespondingauthor{Matthias Fey}{matthias.fey@udo.edu}
\icmlcorrespondingauthor{Jan-Gin Yuen}{jangin.yuen@udo.edu}

\vskip 0.3in
]

\printAffiliationsAndNotice{\icmlEqualContribution}

\newcommand{\ie}{\textit{i.e.}}
\newcommand{\eg}{\textit{e.g.}}
\newcommand{\cf}{\textit{cf.}}
\definecolor{customgray}{rgb}{0.3,0.3,0.3}
\newcommand{\std}[1]{\textcolor{customgray}{\scriptsize{$\pm$#1}}}
\newcommand{\mr}[2]{\multirow{#1}{*}{#2}}
\newcommand{\mc}[3]{\multicolumn{#1}{#2}{#3}}
\newcommand{\mat}[1]{\bm{#1}}
\renewcommand{\vec}[1]{\mathbf{#1}}

\newcommand{\mf}[1]{\textcolor{red}{[Matthias: #1]}}

\definecolor{green}{RGB}{181,208,186}
\definecolor{blue}{RGB}{183,224,238}
\definecolor{orange}{RGB}{254,154,92}

\begin{abstract}
We present a hierarchical neural message passing architecture for learning on molecular graphs. Our model takes in two complementary graph representations: the raw molecular graph representation and its associated junction tree, where nodes represent meaningful clusters in the original graph, \eg, rings or bridged compounds.
We then proceed to learn a molecule's representation by passing messages inside each graph, and exchange messages between the two representations using a coarse-to-fine and fine-to-coarse information flow.
Our method is able to overcome some of the restrictions known from classical GNNs, like detecting cycles, while still being very efficient to train.
We validate its performance on the \textsc{ZINC} dataset and datasets stemming from the \textsc{MoleculeNet} benchmark collection.
\end{abstract}

\section{Introduction}%
\label{sec:introduction}

Machine learning algorithms offer great potentials in reducing the computation time required for predicting molecular properties from several hours to just a few milliseconds \citep{Wu/etal/2018}.
In particular, graph neural networks (GNNs) have been proven to be very successful for this task, exceeding the previously predominated approach of manual feature engineering by a large margin \citep{Gilmer/etal/2017,Schuett/etal/2017}.
In contrast to hand-crafted features, GNNs \emph{learn} high-dimensional embeddings of atoms that are able to represent their complex interactions by exchanging and aggregating messages between them.

In this work, we present a \emph{hierarchical} variant of message passing on molecular graphs.
Here, we utilize two separate graph neural networks that operate on complementary representations of a molecule simultaneously: its raw molecular representation and its corresponding (coarsened) junction tree representation.
Each of the two GNN's intra-message passing step is strengthened by an inter-message passing step that exchanges intermediate information between the two representations.
This allows the network to reason about hierarchy, \eg, rings, in molecules in a natural fashion, and enables the GNN to overcome some of its restrictions, \eg, detecting cycles \citep{Loukas/2020}, without relying on more sophisticated architectures to do so \citep{Morris/etal/2019,Murphy/etal/2019,Maron/etal/2019}.
We show that this simple scheme can drastically increase the performance of a GNN, reaching state-of-the-art performance on a variety of different datasets.
Despite its higher-order nature, our proposed network architecture is still very efficient to train and causes only marginal additional costs in terms of memory and execution time.


\section{Learning on Molecular Graphs}%
\label{sec:learning_on_molecular_graphs}

\emph{Graph neural networks} operate on graph representations of molecules $\mathcal{G} = (\mathcal{V}, \mathcal{E})$, where nodes $\mathcal{V} = \{ 1, \ldots, n \}$ represent atoms and edges $\mathcal{E} \subseteq \mathcal{V} \times \mathcal{V}$ are defined by a predefined structure or by connecting atoms that lie within a certain cutoff distance.
Given atom features $\mat{X}^{(0)} \in \mathbb{R}^{|\mathcal{V}| \times f}$ and edge features $\mat{E}^{(0)} \in \mathbb{R}^{|\mathcal{E}| \times d}$, a GNN iteratively updates node embeddings $\vec{x}_i^{\hspace{-1pt}(l)}$ in layer $l+1$ by aggregating localized information via the parametrized functions
\begin{align*}
  \vec{m}_v^{(l+1)} &= \textsc{Aggregate}^{(l+1)}_{\theta_1} \hspace{-3pt} \left( \hspace{-4pt} {\left\{ \hspace{-5pt} \left\{ \hspace{-2pt} \left( \vec{x}_w^{\hspace{1pt}(l)}, \vec{x}_v^{\hspace{1pt}(l)}, \vec{e}_{w,v}^{\hspace{2pt}(l)} \right) \hspace{-3pt} \right\} \hspace{-5pt} \right\}}_{\hspace{-1pt} w \in \mathcal{N}(v)} \hspace{-1pt} \right) \\
  \vec{x}_v^{\hspace{1pt}(l+1)} &= \textsc{Update}^{(l+1)}_{\theta_2} \hspace{-3pt} \left( \vec{x}_v^{\hspace{1pt}(l)}, \vec{m}_v^{(l+1)} \right) \hspace{-2pt},
\end{align*}
where $\{ \hspace{-3pt} \{ \ldots \} \hspace{-3pt} \}$ denotes a multiset and $\mathcal{N}(v) \subseteq \mathcal{V}$ defines the neighborhood set of node $v \in \mathcal{V}$ \citep{Gilmer/etal/2017}.
After $L$ layers, a graph representation is obtained via global aggregation of $\mat{X}^{(L)}$, \eg, summation or averaging.

Many existing GNNs can be expressed using this neural message passing scheme \citep{Kipf/Welling/2017,Velickovic/etal/2018}.
A GNN called \textsc{GIN} (\textsc{GIN-E} in case edge features are present) \citep{Xu/etal/2019,Hu/etal/2020a}
\begin{equation*}
  \vec{x}_v^{\hspace{1pt}(l+1)} = \textrm{MLP}^{(l+1)}_{\theta} \hspace{-2pt} \left( \hspace{-1pt} (1 + \epsilon) \cdot \vec{x}_v^{\hspace{1pt}(l)} + \hspace{-5pt} \sum_{w \in \mathcal{N}(v)} \hspace{-2pt} \vec{x}_w^{\hspace{1pt}(l)} + \vec{e}_{w,v}^{\hspace{1pt}(l)} \right) \hspace{-3pt},
\end{equation*}
defines its most expressive form, showing high similarities to the popular WL-test \citep{Weisfeiler/Lehman/1968} while being able to operate on continuous node and edge features.

\section{Methodology}%
\label{sec:methodology}

It has been shown that GNNs are unable to distinguish certain molecules when operating on the molecular graph or using limited cutoff distances, \eg, Cyclohexane and two Cyclopropane molecules \citep{Xu/etal/2019,Klicpera/etal/2020}.
These restrictions mostly stem from the fact that GNNs are not capable of detecting cycles \citep{Loukas/2020} since they are unable to maintain information about which vertex in its receptive field has contributed what to the aggregated information \citep{Hy/etal/2018}.
In this section, we present a simple hierarchical scheme to overcome this restriction, which strengthens the GNN's performance with minimal computational overhead in return.

Our method involves learning on two molecular graph representations simultaneously in an end-to-end fashion: the original graph representation and its associated junction tree.
The junction tree representation encodes the tree structure of molecules and defines how clusters (singletons, bonds, rings, bridged compounds) are mutually connected, while the graph structure captures its more fine-grained connectivity \citep{Jin/etal/2018}.
We briefly revisit how junction trees are obtained from molecular graphs before describing our method in detail.

\paragraph{Tree Decomposition.}%
\label{par:tree_decomposition}

\begin{figure}[t]
  \centering
  \includegraphics[width=.8\linewidth]{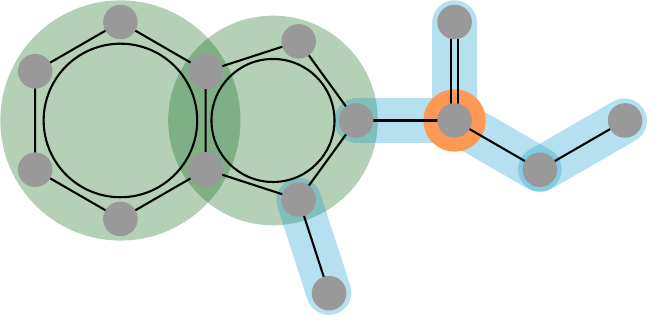}
  \caption{%
    Example of a molecular graph and its cluster assignment for obtaining junction trees.
    Cluster colors refer to \textcolor{orange}{$\blacksquare$} singletons, \textcolor{blue}{$\blacksquare$} bonds and \textcolor{green}{$\blacksquare$} rings.
  }\label{fig:junction}
\end{figure}

Given a graph $\mathcal{G} = (\mathcal{V}, \mathcal{E})$, a \emph{tree decomposition} maps $\mathcal{G}$ into a \emph{junction tree} $\mathcal{T} = (\mathcal{C}, \mathcal{R})$ with node set $\mathcal{C} = \{ \mathcal{C}_1, \ldots, \mathcal{C}_m \}$, $\mathcal{C}_i \subseteq \mathcal{V}$ for all $i \in \{ 1, \ldots, m \}$, and edge set $\mathcal{R} \subseteq \mathcal{C} \times \mathcal{C}$ so that:

\vspace{-0.25cm}

\begin{enumerate}
  \item $\bigcup_{\hspace{1pt}i} \mathcal{C}_i = \mathcal{V}$ and $\bigcup_{\hspace{1pt}i} \mathcal{E}[\mathcal{C}_i] = \mathcal{E}$, where $\mathcal{E}[\mathcal{C}_i] \subseteq \mathcal{C}_i \times \mathcal{C}_i$ represents the edge set of the induced subgraph $\mathcal{G}[\mathcal{C}_i]$
  \item $\mathcal{C}_i \cap \, \mathcal{C}_j \subseteq \mathcal{C}_k$ for all clusters $\mathcal{C}_i$, $\mathcal{C}_k$, $\mathcal{C}_j$ with connections $(\mathcal{C}_i, \mathcal{C}_k) \in \mathcal{R}$ and $(\mathcal{C}_k, \mathcal{C}_j) \in \mathcal{R}$.
\end{enumerate}

\vspace{-0.25cm}

The assignment of atoms $v \in \mathcal{V}$ to clusters $\mathcal{C}_i \in \mathcal{C}$ is given by the matrix $\mat{S} \in {\{ 0, 1 \}}^{|\mathcal{V}| \times |\mathcal{C}|}$ with $S_{v,i} = 1$ iff. $v \in \mathcal{C}_i$.

We closely follow the tree decomposition algorithm of related works \citep{Rarey/Dixon/1998,Jin/etal/2018}.
We first group all simple cycles and all edges that do not belong to any cycle into clusters in $\mathcal{C}$.
Two rings are merged together if they share more than two overlapping atoms (bridged compounds).
For atoms lying inside more than three clusters, we add the intersecting atom as a singleton cluster.
A cluster graph is constructed by adding edges between all intersecting clusters, and the final junction tree $\mathcal{T}$ is then given as one its spanning trees.
Figure~\ref{fig:junction} visualizes how clusters  are formed on an examplary molecule.
For each cluster, we additionally hold its respective category (singleton, bond, ring, bridged compound) as one-hot encodings $\mat{Z}^{(0)} \in {\{0, 1 \}}^{|\mathcal{C}| \times 4}$.

\paragraph{Inter-Message Passing with Junction Trees.}%
\label{par:inter-message_passing_with_junction_trees}

Our method is able to extend any GNN model for molecular property prediction by making use of intra-message passing \emph{in} and inter-message passing \emph{to} a complementary junction tree represention.
Here, instead of using a single GNN operating on the molecular graph, we make use of \emph{two} GNN models: one operating on the original graph $\mathcal{G}$ and one operating on its associated junction tree $\mathcal{T}$, each passing intra-messages to their respective neighbors.
We further enhance this scheme by making use of \emph{inter-message passing}:
Let $\mat{X}^{(l)} \in \mathbb{R}^{|\mathcal{V}| \times h}$ and $\mat{Z}^{(l)} \in \mathbb{R}^{|\mathcal{C}| \times h}$ denote the intermediate representations of $\mathcal{G}$ and $\mathcal{T}$, respectively.
Then, we enhance both representations $\mat{X}^{(l)}$ and $\mat{Z}^{(l)}$ by an additional coarse-to-fine information flow from $\mathcal{T}$ to $\mathcal{G}$
\begin{equation*}
  \mat{X}^{(l)} \leftarrow \mat{X}^{(l)} + \sigma \hspace{-1pt} \left( \mat{S} \mat{Z}^{(l)} \mat{W}_1^{(l)} \right)
\end{equation*}
and reverse fine-to-coarse information flow from $\mathcal{G}$ to $\mathcal{T}$
\begin{equation*}
  \mat{Z}^{(l)} \leftarrow \mat{Z}^{(l)} + \sigma \hspace{-1pt} \left( \mat{S}^{\top} \mat{X}^{(l+1)} \mat{W}_2^{(l)} \right) \hspace{-2pt} ,
\end{equation*}
with $\mat{W}_1, \mat{W}_2 \in \mathbb{R}^{h \times h}$ denoting trainable weights and $\sigma$ being a non-linearity.
This leads to a hierarchical-variant of message passing for learning on molecular graphs, similar to the ones applied in computer vision \citep{Ronneberger/etal/2015,Newell/etal/2016,Lin/etal/2017}.
Furthermore, each atom is able to know about its cluster assignment, and, more importantly, which other nodes are part of the same cluster.
Specifially, this leads to an increased expressivity of GNNs.
For example, the popular example of a Cyclohexane molecule and two Cyclopropane molecules (a single ring and two disconnected rings) \citep{Klicpera/etal/2020} are distinguishable by our scheme since the junction tree representations are distinguishable by the most expressive GNN\@.

\begin{figure*}[t]
  \centering
  \includegraphics[width=.9\linewidth]{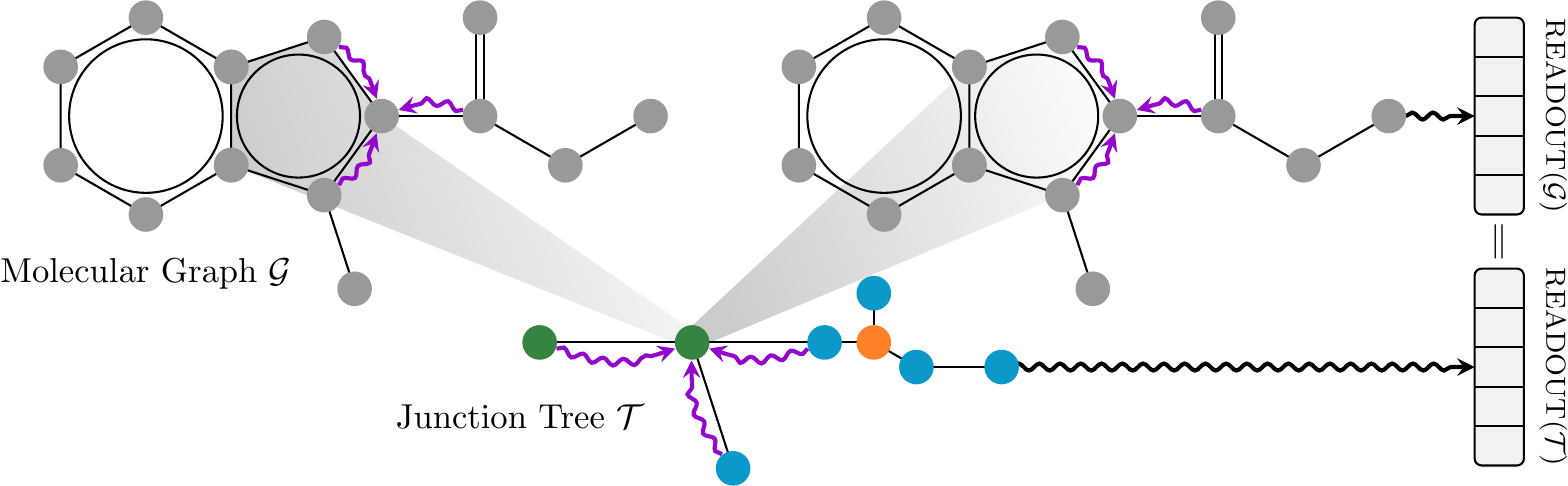}
  \caption{%
    Overview of the proposed approach.
    Two GNNs are operating on the distinct graph representations $\mathcal{G}$ and $\mathcal{T}$, and receive coarse-to-fine and fine-to-coarse information before another round of message passing starts.
    }\label{fig:overview}
\end{figure*}
The \emph{readout} of the model is then given via
\begin{equation*}
  \sum_{v \in \mathcal{V}} \vec{x}^{(L)}_{v} \hspace{1pt} \| \hspace{2pt} \sum_{\mathcal{C}_i \in \mathcal{C}} \vec{z}^{(L)}_{i},
\end{equation*}
with $\|$ denoting the concatenation operator.
A high-level overview of our method is visualized in Figure~\ref{fig:overview}.

\section{Related Work}%
\label{sec:related_work}

We briefly review some of the related work and their relation to our proposed approach.

\paragraph{Learning on Molecular Graphs.}%
\label{par:learning_on_molecular_graphs.}

Instead of using hand-crafted representations \citep{Bartok/etal/2013}, recent advancements in deep graph learning rely on an end-to-end learning of representations which has quickly led to major breakthroughs in machine learning on molecular graphs \citep{Duvenaud/etal/2015,Gilmer/etal/2017,Schuett/etal/2017,Joergensen/etal/2018,Unke/Meuwly/2019,Chen/etal/2019}.
Most of these works are especially designed for learning on the molecular geometry.
Here, earlier models \citep{Schuett/etal/2017,Gilmer/etal/2017,Joergensen/etal/2018,Unke/Meuwly/2019,Chen/etal/2019} fulfill rotational invariance constraints by relying on interatomic distances, while recent models employ more expressive equivariant models.
For example, \textsc{DimeNet} \citep{Klicpera/etal/2020} deploys directional message passing between node triplets to also model angular potentials.
Another line of work breaks symmetries by taking permutations of nodes into account \citep{Murphy/etal/2019,Hy/etal/2018,Albooyeh/etal/2019}.
Recently, it has been shown that strategies for pre-training models on molecular graphs can effectively increase their performance for certain downstream tasks \citep{Hu/etal/2020a}.
Our approach fits nicely into these lines of work since it also increases the expressiveness of GNNs while being orthogonal to further advancements in this field.

\paragraph{Junction Trees.}%
\label{par:junction_trees.}

So far, junction trees have solely been used for molecule generation based on a coarse-to-fine generation procedure \citep{Jin/etal/2018,Jin/etal/2019}.
In contrast to the generation of SMILES strings \citep{Gomez/etal/2018}, this allows the model to enforce chemical validity while generating molecules significantly faster than the node-per-node generation procedure applied in autoregressive methods \citep{You/etal/2018}.

\paragraph{Inter-Message Passing.}%
\label{par:inter-message_passing.}

The idea of inter-message passing between graphs has been already heavily investigated in practice, mostly in the fields of deep graph matching \citep{Wang/etal/2018,Li/etal/2019,Fey/etal/2020} and graph pooling \citep{Ying/etal/2018,Gao/Ji/2019}.
For graph pooling, most works focus on \emph{learning} a coarsened version of the input graph.
However, due to being learned, the coarsened graphs are unable to strengthen the expressiveness of GNNs by design.
For example, \textsc{DiffPool} \citep{Ying/etal/2018} always maps the atoms of two disconnected rings to the \emph{same} cluster, while the $\mathrm{top}_k$ pooling approach \citep{Gao/Ji/2019} either keeps or removes \emph{all} atoms inside those rings (since their node embeddings are shared).
The approach that comes closest to ours involves inter-message passing to a ``virtual'' node that is connected to \emph{all} atoms \citep{Gilmer/etal/2017,Hu/etal/2020b}.
Our approach can be seen as a simple yet effective extension to this procedure.

\section{Experiments}%
\label{sec:experiments}

We evaluate our proposed architecture on the \textsc{ZINC} dataset \citep{Kusner/etal/2017} and a subset of datasets stemming from the \textsc{MoleculeNet} benchmark collection \citep{Wu/etal/2018}.
For all experiments, we make use of the $\textsc{GIN-E}$ operator for learning on the molecular graph \citep{Hu/etal/2020a}, and the $\textsc{GIN}$ operator \citep{Xu/etal/2019} for learning on the associated junction tree.
\textsc{GIN-E} includes edge features (\eg, bond type, bond stereochemistry) by simply adding them to the incoming node features.
All models were trained with the \textsc{Adam} optimizer \citep{Kingma/Ba/2015} using a learning rate of $10^{-4}$, while other hyperparameters (\#epochs, \#layers, hidden size, batch size, dropout ratio) are tuned via an additional validation set.
Our method is implemented in \textsc{PyTorch} \citep{Paszke/etal/2019} and utilizes the \textsc{PyTorch Geometric} \citep{Fey/Lenssen/2019} library.
Our source code is available under \url{https://github.com/rusty1s/himp-gnn}.

\paragraph{ZINC.}%
\label{par:zinc.}

The \textsc{ZINC} dataset \citep{Kusner/etal/2017} contains about $250\,000$ molecular graphs and was introduced in \citet{Dwivedi/etal/2020} as a benchmark for evaluating GNN performances (using a subset of $10\,000$ training graphs).
Here, the task is to regress the constrained solubility of a molecule.
While this is a fairly simple task that can be exactly computed in a short amount of time, it can nonetheless reveal the capabilities across different neural architectures.
We compare ourselves to all the baselines presented in \citet{Dwivedi/etal/2020}, and additionally report results of a $\textsc{GIN-E}$ baseline that does not make use of any additional junction tree information.
Furthermore, we also perform experiments on the full dataset.

\begin{table}[t]
  \centering
  \caption{Results on the \textsc{ZINC} dataset.}\label{tab:zinc}
  \vspace{3pt}
  \begin{tabular}{lcc}
    \mr{2}{\textbf{Method}} & \mc{2}{c}{\textbf{Mean Absolute Error (MAE)}} \\
                            & \textsc{ZINC (10k)} & \textsc{ZINC (Full)} \\
    \toprule
    \textsc{GCN}            & 0.367\std{0.011} & --- \\
    \textsc{GraphSAGE}      & 0.398\std{0.002} & --- \\
    \textsc{GIN}            & 0.408\std{0.008} & --- \\
    \textsc{GAT}            & 0.384\std{0.007} & --- \\
    \textsc{MoNet}          & 0.292\std{0.006} & --- \\
    \textsc{GatedGCN}       & 0.435\std{0.011} & --- \\
    \midrule
    \textsc{GatedGCN-E}     & 0.282\std{0.015} & --- \\
    \textsc{GIN-E}          & 0.252\std{0.014} & 0.088\std{0.002} \\
    \midrule
    \textbf{Ours}           & \textbf{0.151}\std{0.006} & \textbf{0.036}\std{0.002} \\
    \bottomrule
  \end{tabular}
\end{table}

As shown in Table~\ref{tab:zinc}, our method is able to significantly outperform all competing methods.
In comparison to \textsc{GIN-E}, its best perfoming competitor, the additional junction tree extension is able to reduce the error rate about 40--60\%.

\paragraph{MoleculeNet Datasets.}%
\label{par:moleculenet_datasets.}

Following upon \citet{Murphy/etal/2019}, we evaluate our model on the \textsc{HIV}, \textsc{MUV} and \textsc{Tox21} datasets from the \textsc{MoleculeNet} benchmark collection \citep{Wu/etal/2018}, using a 80\%/10\%/10\% random split.
Here, the task is to predict certain molecular properties (cast as binary labels), \eg, whether a molecule inhibits HIV virus replication or not.
We compare ourselves to the neural graph fingerprint (\textsc{NGF}) operator \citep{Duvenaud/etal/2015}, and its relational pooling variant \textsc{RP-NGF} \citep{Murphy/etal/2019}, as well as our own \textsc{GIN-E} baseline.

\begin{table}[t]
  \centering
  \caption{Results on a subset of the \textsc{MoleculeNet} datasets.}\label{tab:moleculenet}
  \vspace{3pt}
  \begin{tabular}{lccc}
    \mr{2}{\textbf{Method}} & \mc{3}{c}{\textbf{ROC-AUC}} \\
                            & \textsc{HIV}    & \textsc{MUV}    & \textsc{Tox21} \\
    \toprule
    \textsc{NGF}            & 81.20\std{1.40} & 79.80\std{2.50} & 79.4\std{1.00} \\
    \textsc{RP-NGF}         & 83.20\std{1.30} & 79.40\std{0.50} & 79.9\std{0.60} \\
    \textsc{GIN-E}          & 83.83\std{0.67} & 79.57\std{1.14} & 86.68\std{0.77} \\
    \midrule
    \textbf{Ours}           & \textbf{84.81}\std{0.42} & \textbf{81.80}\std{2.02} & \textbf{87.36}\std{0.50} \\
    \bottomrule
  \end{tabular}
\end{table}

As the results in Table~\ref{tab:moleculenet} indicate, our method beats both \textsc{NGF} and \textsc{GIN-E} in test performance.
Although \textsc{RP-NGF} is able to distinguish any graph structure by considering permutations of nodes, our approach leads to overall better generalization despite its simplicity.

\paragraph{OGB Datasets.}%
\label{par:ogb_datasets}

We also test the performance of our model on the newly introduced datasets \texttt{ogbg-molhiv} and \texttt{ogbg-molpcba} from the \textsc{OGB} benchmark dataset suite \citep{Hu/etal/2020b}, which are adopted from \textsc{MoleculeNet} and enhanced by a more challenging and standardized scaffold splitting procedure.
We closely follow the experimental protocol of \citet{Hu/etal/2020b} and report ROC-AUC and PRC-AUC for \texttt{ogbg-molhiv} and \texttt{ogbg-molpcba}, respectively.
We compare ourselves to three variants that do not make use of additional junction tree information, namely \textsc{GCN-E}, \textsc{GatedGCN-E} and \textsc{GIN-E} \citep{Kipf/Welling/2017,Bresson/Laurent/2017,Dwivedi/etal/2020,Hu/etal/2020a,Hu/etal/2020b}.

\begin{table}[t]
  \centering
  \caption{Results on the \texttt{molhiv} and \texttt{molpcba} datasets of OGB.}\label{tab:ogbg-mol}
  \vspace{3pt}
  \setlength{\tabcolsep}{5pt}
  \begin{tabular}{lcc}
    \mr{2}{\textbf{Method}} & \textbf{ROC-AUC}         & \textbf{PRC-AUC} \\
                            & \texttt{ogbg-molhiv}     & \texttt{ogbg-molpcba} \\
    \toprule
    \textsc{GCN-E}          & 76.07\std{0.97}          & 19.83\std{0.16} \\
    \textsc{GatedGCN-E}     & 77.65\std{0.50}          & 20.77\std{0.27} \\
    \textsc{GIN-E}          & 75.58\std{1.40}          & 22.17\std{0.23} \\
    \midrule
    \textbf{Ours}           & \textbf{78.80}\std{0.82} & \textbf{27.39}\std{0.17} \\
    \bottomrule
  \end{tabular}
\end{table}

Results are presented in Table~\ref{tab:ogbg-mol}.
As one can see, our approach is able to outperform all its competitors.
Interestingly, our model achieves its best results in combination with a small amount of layers ($2$ or $3$), making its runtime and memory requirements on par with the other baselines (which make use of $5$ layers).
This can be explained by the fact that the additional coarse-to-fine information flow enhances the receptive field size of a GNN, and therefore omits the need to stack a multitude of layers.

\section{Conclusion}%
\label{sec:conclusion}

We introduced an end-to-end architecture for molecular property prediction that utilizes inter-message passing between graph representations of different hierarchy.
Our proposed method can be used as a plug-and-play extension to strengthen the capabilities of a GNN operating on molecular graphs with little to no overhead.
In future works, we are interested in studying how the proposed approach can be applied to other domains as well, \eg, social networks.



\bibliography{main}
\bibliographystyle{icml2020}

\end{document}